\documentclass[letterpaper, 10 pt, conference]{ieeeconf}  

\IEEEoverridecommandlockouts                              

\usepackage{epsfig} 
\usepackage{mathptmx} 
\usepackage{times} 
\usepackage{amsmath} 
\usepackage{amssymb}  
\usepackage[hidelinks]{hyperref}
\usepackage{amsmath}
\usepackage{amsfonts}
\usepackage{amssymb}
\usepackage{graphicx}
\usepackage{xcolor}
\usepackage{physics}
\usepackage{multicol}
\graphicspath{{figures/}}
\usepackage{mathtools}
\usepackage{caption}
\usepackage{subcaption}
\usepackage{graphicx}
\usepackage{wrapfig}
\usepackage[export]{adjustbox}
\usepackage{physics}
\usepackage{siunitx}
\usepackage{comment}
\usepackage{float}
\usepackage{cleveref}
\usepackage{pgfplots}
\usepackage{tikzscale}
\usepackage{booktabs}
\usepackage{cite}

\newcommand{\R}{\mathbb{R}}

\definecolor{julia_green}{rgb}{0.22, 0.596, 0.149}

\title{\LARGE \bf
Aquarium: A Fully Differentiable Fluid-Structure Interaction Solver for Robotics Applications
}

\author{Jeong Hun Lee$^{1}$, Mike Y. Michelis$^{2}$, Robert Katzschmann$^{2}$, and Zachary Manchester$^{1}$
\thanks{*This material is based upon work supported by the National Science Foundation Graduate Research Fellowship under Grant No. DGE2140739.}
\thanks{$^{1}$ The Robotics Institute, Carnegie Mellon University, Pittsburgh, USA
        {\tt\small jeonghunlee@cmu.edu, zacm@cmu.edu}}%
\thanks{$^{2}$ Soft Robotics Lab, ETH Zurich, Zurich, Switzerland
        {\tt\small michelism@ethz.ch, rkk@ethz.ch}}%
}

\begin{document}

\maketitle
\thispagestyle{empty}
\pagestyle{empty}

\begin{abstract}

We present Aquarium, a differentiable fluid-structure interaction solver for robotics that offers stable simulation, accurately coupled fluid-robot physics in two dimensions, and full differentiability with respect to fluid and robot states and parameters. Aquarium achieves stable simulation with accurate flow physics by directly integrating over the incompressible Navier-Stokes equations using a fully implicit Crank-Nicolson scheme with a second-order finite-volume spatial discretization. The fluid and robot physics are coupled using the immersed-boundary method by formulating the no-slip condition as an equality constraint applied directly to the Navier-Stokes system. This choice of coupling allows the fluid-structure interaction to be posed and solved as a nonlinear optimization problem. This optimization-based formulation is then exploited using the implicit-function theorem to compute derivatives. Derivatives can then be passed to downstream gradient-based optimization or learning algorithms. We demonstrate Aquarium's ability to accurately simulate coupled fluid-robot physics with numerous 2D examples, including a cylinder in free stream and a soft robotic fish tail with hardware validation. We also demonstrate Aquarium's ability to provide analytical gradients by performing gradient-based shape-and-gait optimization of an oscillating diamond foil to maximize its generated thrust.

\end{abstract}


\begin{figure}[t!]
    \vspace{0.25\baselineskip}
    \centering
    \begin{subfigure}[t]{0.48\textwidth}
        \includegraphics[width=\textwidth, trim={0 15cm 700 0},clip]{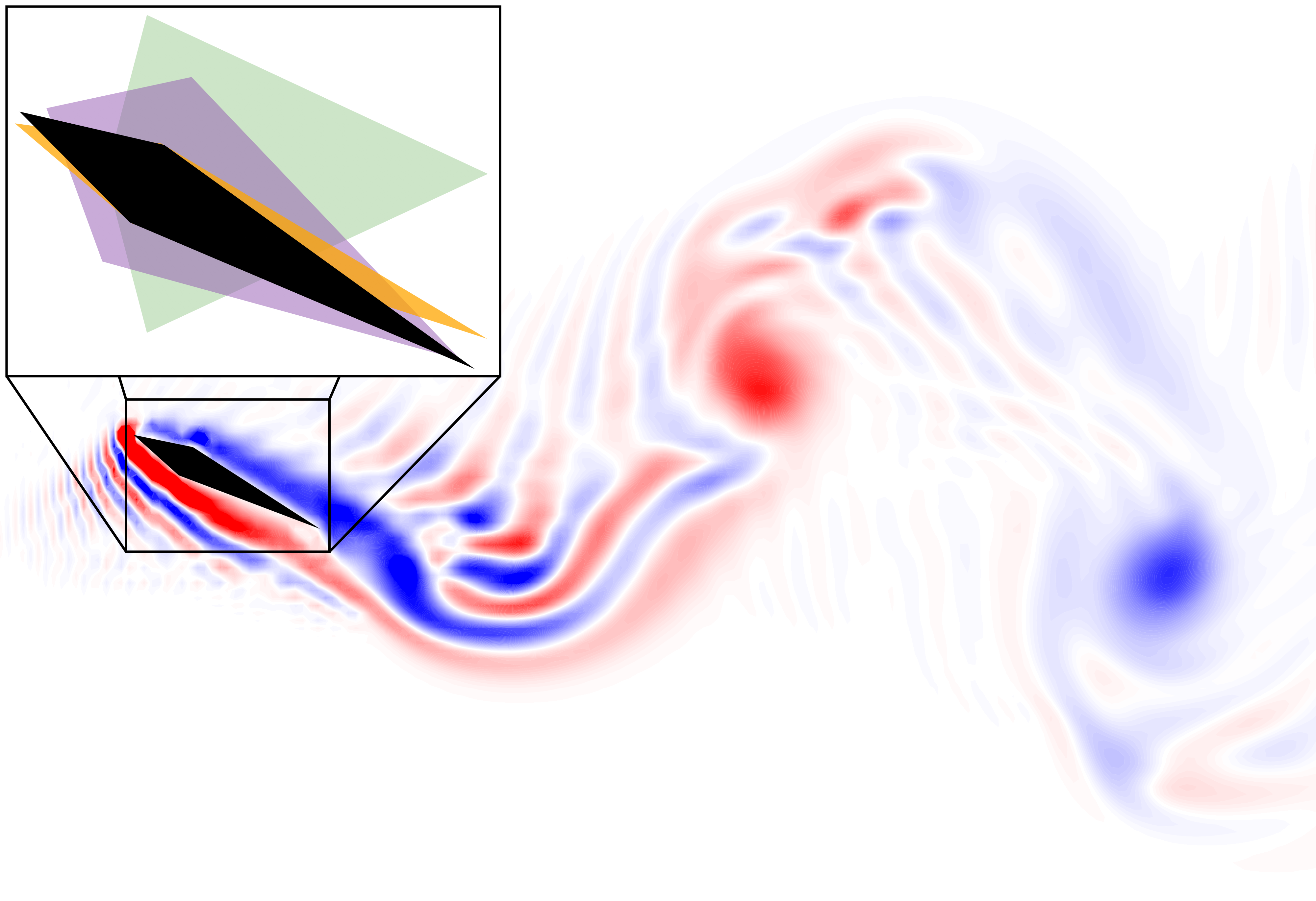}
        \caption{Progression of foil shape-gait profiles throughout the optimization process. Shape colors are in reference to those in Figure~\ref{fig:optimization-thrust-history}.}
        \label{fig:optimization-shape-progression}
    \end{subfigure}
    \vspace{0.5\baselineskip}
    \\
    \begin{subfigure}[t]{0.48\textwidth}
        \includegraphics[width =\textwidth, height = 0.6165\textwidth]{BFGS_thrust_time_history.tikz}
        \caption{Time history of generated thrust by foil shape-gait profiles throughout the optimization process.}
        \label{fig:optimization-thrust-history}
    \end{subfigure}
    \caption{Shape-and-gait co-optimization of an oscillating diamond foil using the limited-memory BFGS (L-BFGS) algorithm with analytical gradients provided by Aquarium. The foil thickness, heave, pitch, and pitch-heave phase are optimized to maximize the generated thrust. A fully converged solution (black) is found after only 14 L-BFGS iterations from a drag-inducing initial guess (\color{julia_green}green\color{black}).}
    \label{fig:foil-optimization}
    \vspace{-\baselineskip}
\end{figure}

\section{INTRODUCTION} \label{sec:INTRO}

In recent years, there has been considerable interest in bio-inspired locomotion for underwater~\cite{maertens_optimal_2017, gao_independent_2018, novati_synchronisation_2017, katzschmann_exploration_2018, miles_dont_2019, jiao_learning_2021} and aerial vehicles~\cite{platzer_flapping_2008, shyy_recent_2010, loquercio_deep_2019, appius_raptor_2022} involving complex interactions with the fluid environment. For example, the energy efficiency and high maneuverability of fish-like propulsion are attributed to the sensing and shedding of vortices~\cite{triantafyllou_efficient_1995, triantafyllou_hydrodynamics_2000}. This complex fluid interaction is also present in the flight of flapping-wing aerial vehicles~\cite{shyy_flapping_2007} and high-angle-of-attack perching maneuvers of fixed-wing airplanes~\cite{manchester_variable_2017, moore_robust_2014}, and has led to the development of various robotic hardware systems, including oscillating-foil propulsion for marine vehicles~\cite{anderson_oscillating_1998}, various biomimetic underwater systems~\cite{katzschmann_exploration_2018, anderson_maneuvering_2002, christianson_cephalopod-inspired_2020, aubin_electrolytic_2019}, and flapping-wing, micro-aerial vehicles~\cite{chen_biologically_2017, jafferis_untethered_2019, farrell_helbling_review_2018}.

In contrast, there has been comparatively little work \emph{jointly} optimizing both the systems' design and control parameters~\cite{zhang_learning_2021, ma_diffaqua_2021} while reasoning about the fluid environment~\cite{van_rees_optimal_2015, nava_fast_2022}. Such an optimization could provide insight into the well-known passive swimming dynamics of a dead trout~\cite{beal_passive_2006} and lead to robotic systems with similar capabilities. We believe that a major reason for the lack of work in this direction is the absence of an open-source, accurate, stable simulator with~\emph{full differentiability}.

In recent years, various differentiable simulators~\cite{howell_dojo_2022, tedrake_drake_2019, todorov_mujoco_2012, heiden_neuralsim_2021, freeman_braxdifferentiable_2021, geilinger_add_2020, werling_fast_2021, degrave_differentiable_2019, du_diffpd_2021, hu_chainqueen_2019} have been deployed for a wide range of uses in robotics. These solvers are able to provide gradients that can be passed directly to optimization-based frameworks such as reinforcement learning and model-predictive control. Differentiable simulators can also be integrated into neural networks during backpropogation to achieve high sample efficiency and accuracy~\cite{ma_diffaqua_2021, amos_optnet_2017, de_avila_belbute-peres_end--end_2018}.

In the fluids community, various computational fluid dynamics (CFD) simulators also exist~\cite{nava_fast_2022, chen_computational_2014, battista_ib2d_2017, bernier_simulations_2019, jasak_openfoam_2009, palacios_stanford_2014, manual_ansys_2009, siemens_digital_industries_software_simcenter_2021,  holl_phiflow_2020, bezgin_jax-fluids_2022, du_functional_2020, liu_fishgym_2022}. However, these solvers each possess at least one key deficiency that limits their usability for robotics. These deficiencies include poor accuracy, generalizability, and computational efficiency; instability over larger time steps due to explicit integration; lack of full differentiability for robotics tasks (e.g., control) in unsteady flow; and inability to handle fluid-structure interaction (FSI) to properly simulate the unsteady, multi-physics coupling between the fluid and robot dynamics.

We propose \textbf{Aquarium}, an \emph{open-source, physics-based, fully differentiable} solver for simulating the two-dimensional (2D) coupled dynamics between robotic systems and their surrounding fluid environment. The single-phase fluid dynamics are solved by integrating over the governing Navier-Stokes equations directly with a fully implicit Crank-Nicolson scheme to preserve stability over large time steps. The fluid discretization is handled using a second-order finite-volume method, while the fluid-robot coupling is achieved using the immersed boundary method~\cite{peskin_immersed_2002}, which separates the fluid and robot meshes to avoid computationally expensive re-meshing. Specifically, we build upon the work of Taira et. al. \cite{taira_immersed_2007} and Perot \cite{perot_analysis_1993} to pose the FSI dynamics as an optimization problem, with the fluid-robot coupling acting as equality constraints to satisfy the no-slip boundary condition. Analytical gradients are computed by applying the implicit function theorem directly to the FSI problem. Aquarium is currently implemented in 2D for rigid bodies, but the methodology is generalizable to 3D flow and soft bodies. In summary, Aquarium offers:

\begin{itemize}

  \item Simulations that solve the discretized 2D Navier-Stokes equations directly with multi-physics coupling between rigid bodies and a fluid environment.
  
  \item Fully implicit time integration using Crank-Nicholson to achieve stable simulation at reasonable sample rates for control and optimization.
  
  \item Full differentiability to calculate analytical gradients with respect to fluid and robot states and parameters for use in gradient-based optimization and learning frameworks.
  
\end{itemize}

The remainder of the paper is organized as follows: In \Cref{sec:RELATED WORKS}, we provide some background on existing CFD and FSI solvers, including their uses and limitations. \Cref{sec:METHODOLOGY} then describes the proposed Aquarium solver. In \Cref{sec:RESULTS}, we provide simulation results and hardware validation on a variety of examples, including a cylinder in free stream and a flapping, soft robotic fish tail in initially still water. We then showcase the differentiability of Aquarium by performing gradient-based shape-and-gait co-optimization of an oscillating diamond foil to maximize its generated thrust. In \Cref{sec:CONCLUSION} we provide final concluding remarks and discuss future work to address current limitations.

\section{RELATED WORKS} \label{sec:RELATED WORKS}

\subsection{Differentiable Fluid Dynamics}

Industry-standard CFD solvers such as OpenFOAM~\cite{jasak_openfoam_2009}, SU2~\cite{palacios_stanford_2014}, ANSYS FLUENT~\cite{manual_ansys_2009}, and STAR-CCM+~\cite{siemens_digital_industries_software_simcenter_2021} provide accurate results for complex flows (e.g., multi-phase, heat transfer, etc.) but at high computational costs with only OpenFOAM and SU2 being open-source. Consequently, running parameter sweeps for gradient-free optimization can quickly become prohibitively expensive for high-dimensional problems. In addition, gradient-free approaches tend to be less stable and slow to convergence when compared to their gradient-based counterparts~\cite{zhang_learning_2021, du_diffpd_2021}.

The class of differentiable simulators addresses this shortcoming. Though many differentiable solvers were developed for robotics~\cite{howell_dojo_2022, tedrake_drake_2019, todorov_mujoco_2012, heiden_neuralsim_2021, freeman_braxdifferentiable_2021, geilinger_add_2020, werling_fast_2021, degrave_differentiable_2019, du_diffpd_2021, hu_chainqueen_2019}, there are also several works aimed towards the fluids community. PhiFlow~\cite{holl_phiflow_2020} was developed as a differentiable PDE solver for deep-learning, written in frameworks that allow for automatic differentiation, such as JAX~\cite{bradbury_jax_2018}, PyTorch~\cite{paszke_pytorch_2019}, and TensorFlow~\cite{martin_abadi_tensorflow_2015}. However, this method is mainly directed at controlling fluids directly by solving the governing Navier-Stokes equations and does not support FSI. Similarly, JAX-FLUIDS~\cite{bezgin_jax-fluids_2022} developed a level-set method for differentiable, compressible, two-phase fluid simulations in JAX. The work implements various boundary conditions, including immersed boundaries for rigid bodies, but only supports explicit integration and, similar to PhiFlow, focuses on deep learning of fluid dynamics and does not currently support FSI.

\subsection{Design Optimization using Adjoint Methods}

While not usually called ``differentiable simulators,'' several conventional CFD frameworks~\cite{jasak_openfoam_2009, palacios_stanford_2014, siemens_digital_industries_software_simcenter_2021, manual_ansys_2009} implement adjoint methods~\cite{jameson_aerodynamic_1988, kenway_effective_2019} to efficiently provide gradients for large-scale shape optimization problems in steady-state flow conditions. SU2 additionally offers shape optimization capabilities in unsteady environments. Rather than solving for the fluid-model derivatives explicitly, these frameworks calculate Jacobian-vector products to efficiently calculate gradients of the optimization problem. This is equivalent to reverse-mode automatic differentiation that is widely used in machine learning~\cite{bradbury_jax_2018, paszke_pytorch_2019, martin_abadi_tensorflow_2015}. However, the extension of adjoint methods to other applications that are critical for robotics (e.g., control in unsteady flow) are relatively unexplored and not available as open-source platforms~\cite{bazilevs_adjoint-based_2013, rumpfkeil_optimal_2010, an_optimal_2021}.
 
\subsection{Fluid-Structure Interaction for Optimization}

As previously mentioned, conventional CFD simulators~\cite{jasak_openfoam_2009, palacios_stanford_2014, siemens_digital_industries_software_simcenter_2021, manual_ansys_2009} are currently limited to gradient-based shape optimization via adjoint methods. Using FSI for gradient-based, non-shape (e.g., gait) optimization in unsteady flow is still challenging, with previous work simplifying the fluid model to potential flow~\cite{jiao_learning_2021} or Stokes flow~\cite{du_functional_2020, grover_geometric_2018, grover_motion_2019} for low-Reynolds-number regimes. Recently, Nava et al. proposed a physics-informed, neural-network model of FSI for the optimization of soft robotic swimmers~\cite{nava_fast_2022}. However, no guarantees can be given when generalizing to new shapes and flow conditions.

Finally, Liu et al. \cite{liu_fishgym_2022} have implemented an FSI extension based in the OpenAI Gym environment. While not differentiable, their fluid solver implementation, a GPU-optimized lattice-Boltzmann method, is highly efficient and can be integrated with reinforcement-learning pipelines. The FSI is achieved with an immersed-boundary method and various swimming bodies are modeled as articulated rigid bodies. The advantages of coupling robot dynamics with full fluid solvers, as opposed to approximated fluid models \cite{manchester_variable_2017, tu_artificial_1994, min_softcon_2019} are shown in several applications, such as leveraging K\'arm\'an vortices for faster propulsion in swimming~\cite{van_rees_optimal_2015, maertens_optimal_2017} and greater lift in flight of flapping-wing microrobots~\cite{chen_computational_2014}.

\section{DIFFERENTIABLE FLUID-STRUCTURE INTERACTION} \label{sec:METHODOLOGY}

Computationally modeling fluid dynamics and fluid-structure interaction using the Navier-Stokes equations has been extensively studied~\cite{peskin_immersed_2002, taira_immersed_2007, perot_analysis_1993, lai_immersed_2000, kim_application_1985, patankar_calculation_1983}. Rather than describing the methods in detail, we highlight the key concepts and refer the reader to existing literature on CFD and FSI for more details. Specifically, our work is most closely related to that of Taira et. al~\cite{taira_immersed_2007} and Perot~\cite{perot_analysis_1993}.

\subsection{Implicit Fluid Model}

We begin with the non-dimensionalized, incompressible Navier-Stokes equations, which express conservation of momentum and mass for Newtonian fluids:
\begin{align} 
    \frac{du}{dt} + (u \cdot \nabla)u & = - \nabla p + \frac{1}{Re}\nabla^{2}u + a_{ext}, \label{eq:Navier-Stokes-momentum} \\
    \nabla \cdot u & = 0, \label{eq:Navier-Stokes-continuity}
\end{align}
where $u$, $p$, $a_{ext}$, and $Re$ are the fluid velocities, pressure, acceleration due to external forces (i.e., gravity, etc.), and non-dimensional Reynolds number, respectively. The Reynolds number is further defined as
\begin{equation}
    Re = \frac{\rho u_{ref}l_{ref}}{\mu},
\end{equation}
where $\rho$ is the fluid density, $\mu$ is the dynamic viscosity of the fluid, $u_{ref}$ is a reference velocity (e.g., free-stream velocity), and $l_{ref}$ is a reference length (e.g., width of the robot).

Using a second-order finite-volume method, we express the continuous, partial derivatives of \eqref{eq:Navier-Stokes-momentum} and \eqref{eq:Navier-Stokes-continuity} as discrete operations over a spatial fluid grid:
\begin{multicols}{2}
\noindent
    \begin{align*}
        (u \cdot \nabla)u & \Rightarrow N(u), \\
        \nabla p & \Rightarrow Gp,
    \end{align*}
    \begin{align*}
        \nabla^{2}u & \Rightarrow Lu + bc_{L}, \\
        \nabla \cdot u & \Rightarrow Du + bc_{D},
    \end{align*}
\end{multicols}
\noindent where $N(u) \in \R^{n_{u}}$ is the nonlinear convective term, $G \in \R^{n_{u} \times n_{p}}$ is the gradient operator, $L \in \R^{n_{u} \times n_{u}}$ is the Laplacian operator, $D \in \R^{n_{p} \times n_{u}}$ is the divergence operator, $bc_{L} \in \R^{n_{u}}$ is the boundary condition term corresponding to the Laplacian, and $bc_{D} \in \R^{n_{p}}$ is the boundary condition term corresponding to the divergence. Then, using implicit Crank-Nicolson integration over a time step, we have the discrete, incompressible Navier-Stokes equations,
\begin{align}
    R(u_{k+1}, u_{k}, p_{k+1}) & = \begin{multlined}[t] Au_{k+1} + \dfrac{1}{2}N(u_{k+1}) - {} \\
        r(u_{k}) + Gp_{k+1} = 0, \label{eq:Navier-Stokes-momentum-discrete}
    \end{multlined} \\
    c_{1}(u_{k+1}) & = Du_{k+1} + bc_{D} = 0, \label{eq:Navier-Stokes-continuity-discrete}
\end{align}
where $u_{k+1} = [u_{x}, u_{y}]^{T}_{k+1} \in \R^{n_{u}}$, $p_{k+1} \in \R^{n_{p}}$, and $A$ and $r$ are defined as follows:
\begin{align}
    A & = \frac{1}{\Delta t}I - \frac{1}{2Re}L, \\
    r(u_{k}) & = \Bigr[ \frac{1}{\Delta t}I + \frac{1}{2Re}L \Bigr] u_{k} - \frac{1}{2}N(u_{k}) + \frac{1}{Re}bc_{L} + a_{ext}.
\end{align}
Interestingly, it has been noted that $-G^{T} = D$ \cite{taira_immersed_2007}. Therefore, $p_{k+1}$ can be interpreted as a Lagrange multiplier enforcing the conservation-of-mass constraint imposed by \eqref{eq:Navier-Stokes-continuity-discrete}.

\subsection{Fluid-Structure Interaction Model} 

\begin{figure}[t]
    \vspace{0.5\baselineskip}
    \centering
    \includegraphics[width=0.45\textwidth]{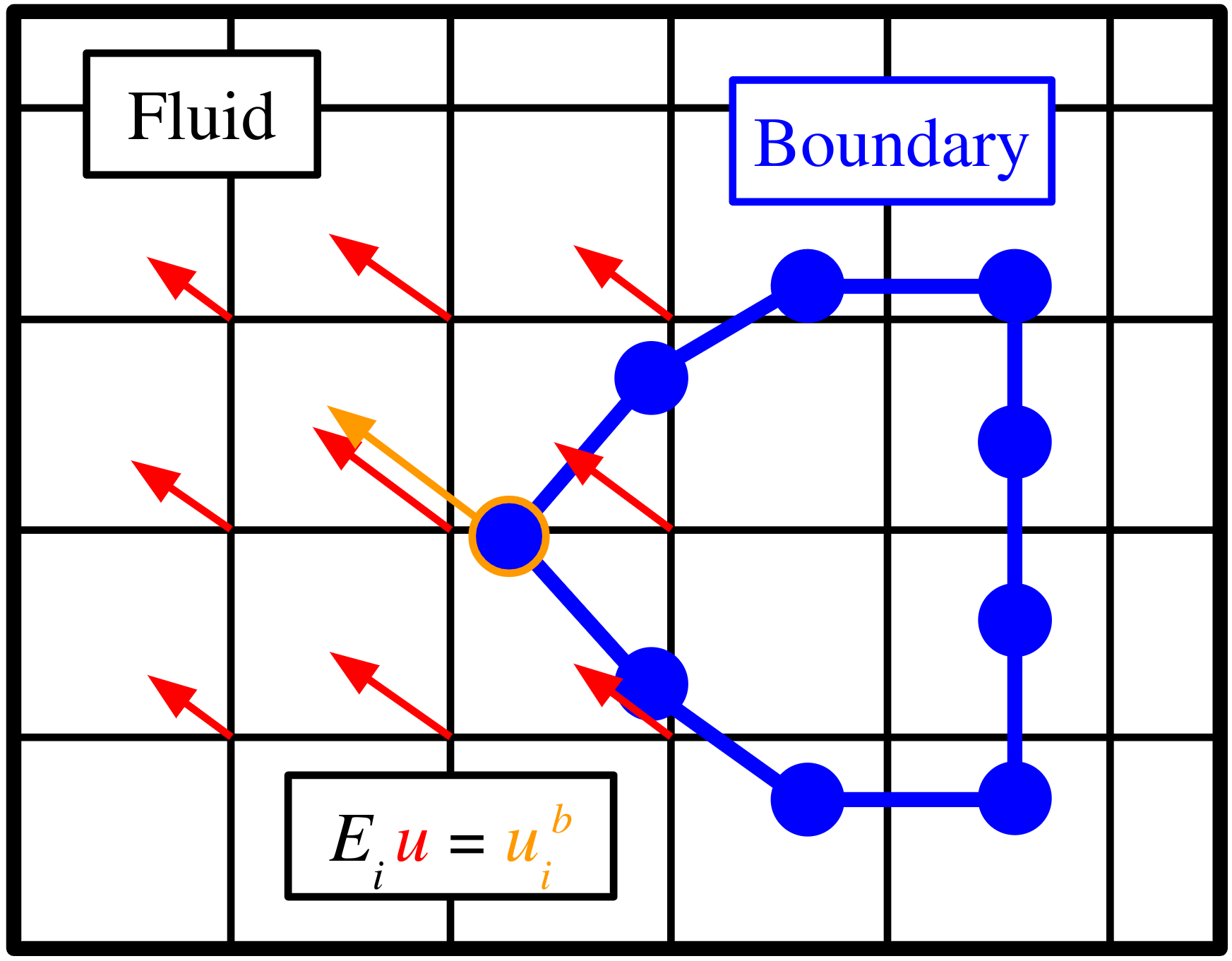}
    \caption{The immersed-boundary method, where the fluid domain is represented by a fixed Eulerian grid (black), and the boundary of the rigid body (i.e., robot) is represented by a moving Lagrangian mesh (\color{blue}blue\color{black}). The meshes are coupled by a convolution matrix $E$ that maps fluid-cell velocities (\color{red}red\color{black}) to those of the boundary nodes (\color{orange}orange\color{black}).}
    \label{fig:immersed_boundary_visual}
    \vspace{-\baselineskip}
\end{figure}

To model the FSI, we extend the idea of treating the Navier-Stokes equations as an optimization problem. First, the FSI is modeled as an additional constraint in the form of no-slip boundary conditions, where the velocity of the fluid must equal the velocity of the robot at the boundary. To do so, we use the immersed-boundary method \cite{peskin_immersed_2002}, which separates the fluid ($u$) and robot ($x^{b}$) meshes into a fixed Eulerian grid and a floating Lagrangian-boundary mesh, respectively, as illustrated in Figure \ref{fig:immersed_boundary_visual}. The coupling between the meshes is formulated as a convolution matrix, $E \in \R^{n_{b} \times n_{u}}$ that maps the fluid cell velocities to the boundary node locations, allowing us to formulate the no-slip constraint,
\begin{equation}
    E(\theta)u_{k+1} = u^{b}_{k+1}, \label{eq:no-slip-condition}
\end{equation}
where $u^{b}_{k+1} \in \R^{n_{b}}$ contains the velocities of the boundary nodes and $\theta \in \R^{n_{\theta}}$ represents parameters of the robot, such as shape parameters and the state, $x_{k+1}$. We then apply \eqref{eq:no-slip-condition} to our Navier-Stokes formulation as an additional constraint, providing us with the full FSI problem:
\begin{align}
    R \left( \begin{multlined} \resizebox{.22\hsize}{!}{$u_{k+1}, u_{k}, p_{k+1},$} \\
    \resizebox{.11\hsize}{!}{$\Tilde{f}^{b}_{k+1}, \theta$} \end{multlined}\right) = & \begin{multlined}  \resizebox{.45\hsize}{!}{$Au_{k+1} + \dfrac{1}{2}N(u_{k+1}) - r(u_{k})$} {}\\
     \resizebox{.39\hsize}{!}{$+ Gp_{k+1} + E(\theta)^{T}\Tilde{f}^{b}_{k+1} = 0$},
    \end{multlined} \label{eq:FSI-momentum-discrete} \\
    c_{1}(u_{k+1}) = & G^{T}u_{k+1} - bc_{D} = 0, \label{eq:FSI-continuity-discrete} \\
    c_{2}(u_{k+1}, \theta) = & E(\theta)u_{k+1} - u^{b}_{k+1} = 0. \label{eq:FSI-continuity-discrete}
\end{align}

We solve \eqref{eq:FSI-momentum-discrete}--\eqref{eq:FSI-continuity-discrete} using a Gauss-Newton method, in which the equations are locally linearized to compute an update step,
\begin{align}
    \begin{bmatrix}
        A + \frac{1}{2}\frac{\partial N}{\partial u_{k+1}} & G & E^{T}\\
        G^{T} & 0 & 0 \\
        E & 0 & 0 \\
    \end{bmatrix} \begin{bmatrix}
        \Delta u_{k+1}\\
        \Delta p_{k+1}\\
        \Delta \Tilde{f}^{b}_{k+1}\\
    \end{bmatrix} = \begin{bmatrix}
        -R\\
        -c_{1}\\
        -c_{2}\\
    \end{bmatrix} \label{eq:FSI-KKT}
\end{align}
where $\Tilde{f}^{b}_{k+1} \in \R^{n_{b}}$ acts as the dual variable for \eqref{eq:no-slip-condition}. Equation \eqref{eq:FSI-KKT} has the structure of a Karush-Kuhn-Tucker (KKT) system, which are common in constrained optimization~\cite{nocedal_numerical_2006}. Upon inspection, $\Tilde{f}^{b}_{k+1}$ effectively acts as a non-dimensional acceleration that fluid particles experience at the boundary. Therefore, forces acting on each boundary node (i.e., pressure) can be calculated directly as
\begin{equation}
    f^{b}_{k+1} = -\rho \frac{h_{x}h_{y}}{s}\Tilde{f}^{b}_{k+1}, \label{eq:boundary-pressure}
\end{equation}
where $f^{b}_{k+1} \in \R^{n_{b}}$ are the pressure forces acting along the boundary, $\rho$ is the fluid density, $h_{x}$ and $h_{y}$ are the spatial step sizes of the fluid grid discretization, and $s$ is the step size of the boundary discretization. $f^{b}_{k+1}$ can then be integrated along the surface of the robot body to obtain net forces.

\subsection{Simulation Gradients}

We exploit the structure of the KKT system defined by \eqref{eq:FSI-KKT} to calculate analytical Jacobians of our FSI model \cite{barratt_differentiability_2018}. For clarity, we first start by looking at our system at time $t_{k}$ and group \eqref{eq:FSI-momentum-discrete}--\eqref{eq:FSI-continuity-discrete} to simplify the model:
\begin{align}
    z & = (u_{k+1}, u_{k}, p_{k+1}, \Tilde{f}^{b}_{k+1}) \\
    g(z; \theta) & = \begin{bmatrix}
        R(u_{k+1}, u_{k}, p_{k+1}, \Tilde{f}^{b}_{k+1}, \theta) \\
        c_{1}(u_{k+1}) \\
        c_{2}(u_{k+1}, \theta) \\
    \end{bmatrix} = 0. \label{eq:simple-KKT}
\end{align}
By definition, $g(z^{*}; \theta) = 0$ is an implicit function, where $z^{*}$ represents the converged solution at each time-step. Using the implicit function theorem \cite{dini_lezioni_1907}, we then compute the derivative $\frac{\partial z}{\partial \theta}$:
\begin{equation}
    \frac{\partial g}{\partial z} \delta z + \frac{\partial g}{\partial \theta} \delta \theta = 0 \\ \implies \frac{\partial z}{\partial \theta} = -\left( \frac{\partial g}{\partial z} \right)^{-1} \frac{\partial g}{\partial \theta}. \label{eq:IFT-definition}
\end{equation} 
Expanding \eqref{eq:IFT-definition}, we arrive at
\begin{equation}
    \resizebox{.88\hsize}{!}{$-\underbrace{\begin{bmatrix}
        A + \frac{1}{2}\frac{\partial N}{\partial u_{k+1}} & G & E^{T}\\
        G^{T} & 0 & 0 \\
        E & 0 & 0 \\
    \end{bmatrix}}_{D_{g}'} \begin{bmatrix}
        \frac{\partial u_{k+1}}{\partial \theta}\\
        \frac{\partial p_{k+1}}{\partial \theta}\\
        \frac{\partial \Tilde{f}^{b}_{k+1}}{\partial \theta}\\
    \end{bmatrix} = \begin{bmatrix}
        -\frac{\partial r}{\partial u_{k}} \\ 0 \\ 0
    \end{bmatrix} \frac{\partial u_{k}}{\partial \theta} + \frac{\partial g}{\partial \theta}$}, \label{eq:gradient-KKT-system}
\end{equation}
 where $D_{g}'$ is the same KKT system that appears in \eqref{eq:FSI-KKT} and $\frac{\partial u_{k}}{\partial \theta}$ has already been computed at the previous time step $t_{k-1}$. Therefore, we can re-use the matrix factorization computed during the simulation step to calculate derivatives for very little additional computational cost. These Jacobians can then be passed to gradient-based solvers to optimize shapes, gaits, controls, or trajectories. This method also generalizes to gradient computations with respect to fluid states and parameters (e.g., fluid density).
\begin{figure}[b]
    \vspace{-\baselineskip}
    \centering
    \includegraphics[width=0.47\textwidth]{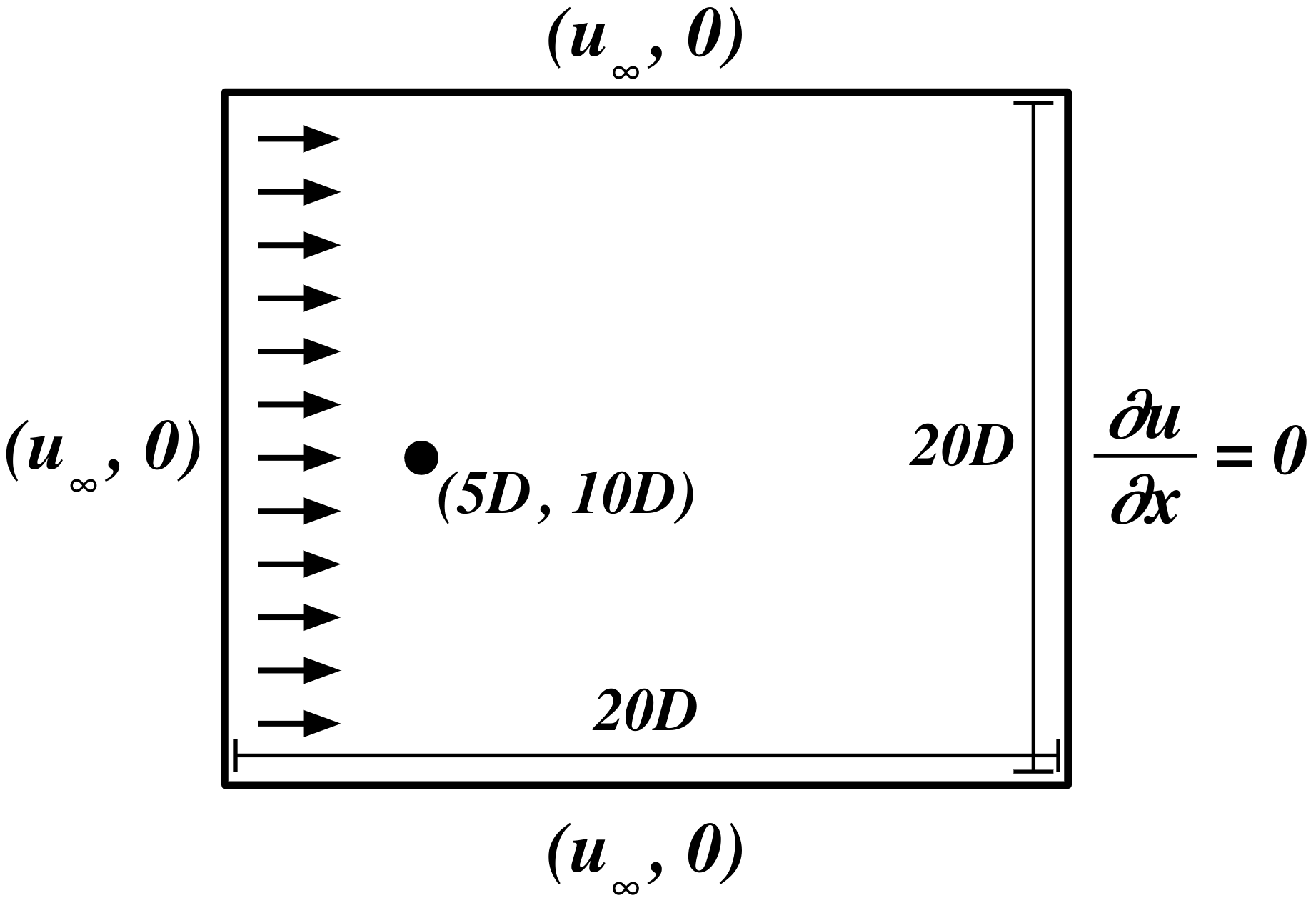}
    \caption{Cylinder-in-free-stream simulation setup with inflow and outflow boundary conditions defined for left and right boundaries, respectively. Far-field boundary conditions are defined for the top and bottom boundaries.}
    \vspace{-0.3\baselineskip}
    \label{fig:cylinder_setup}
\end{figure}
\begin{figure*}[t] \centering
    \vspace{0.5\baselineskip}
    \begin{subfigure}[t]{0.49\textwidth}
        \includegraphics[width=\textwidth]{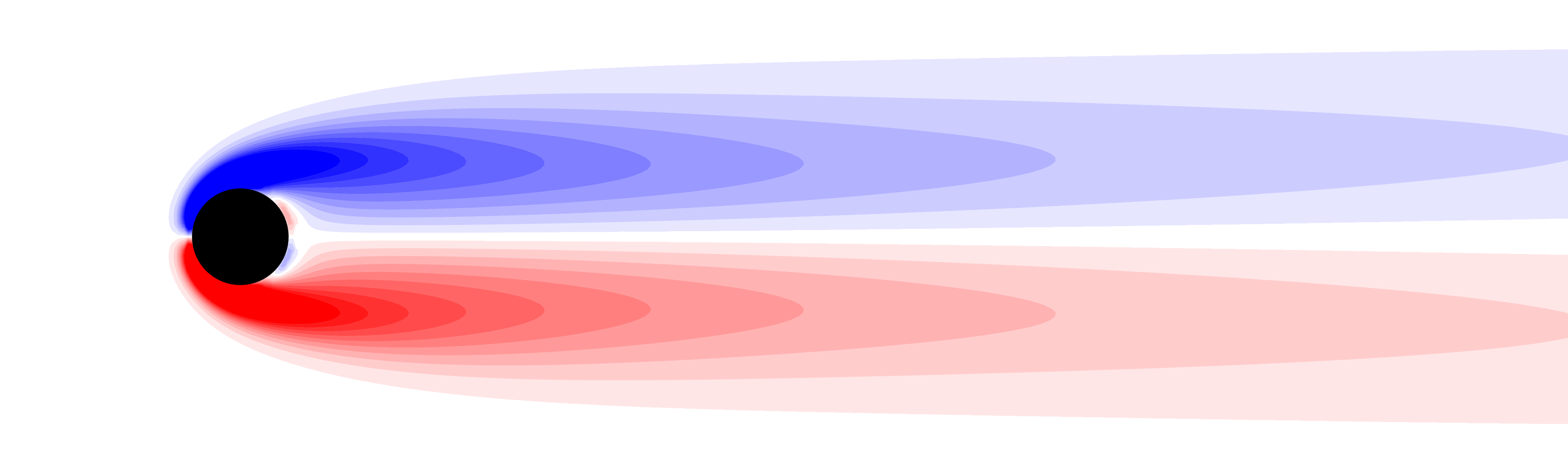}
        \caption{$Re=40$}
        \label{fig:freestream-cylinder-Re40}
    \end{subfigure}
    \begin{subfigure}[t]{0.49\textwidth}
        \includegraphics[width=\textwidth]{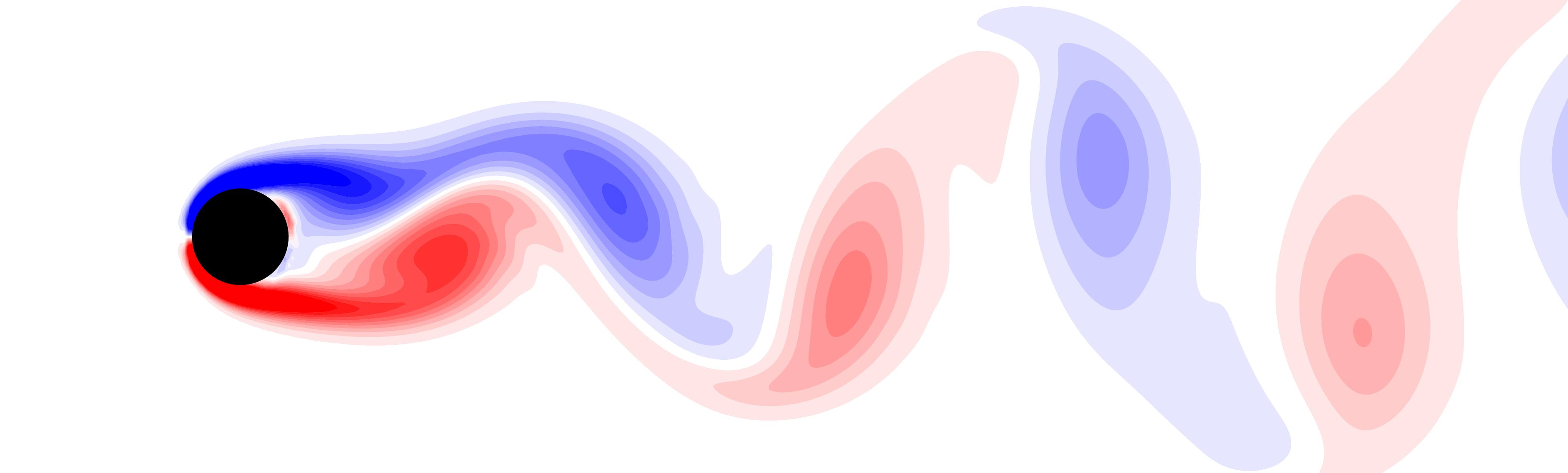}
        \caption{$Re=100$}
        \label{fig:freestream-cylinder-Re100}
    \end{subfigure}
    \caption{Vorticity contours of flow around a cylinder in steady-state, free-stream conditions at varying Reynolds numbers. Being Navier-Stokes based, Aquarium is able to properly simulate the vortex-shedding that occurs at higher Reynolds numbers.}
    \label{fig:freestream-cylinder-vorticity}
\end{figure*}
\begin{figure*}[t] \centering
    \vspace{-0.25\baselineskip}
    \begin{subfigure}[t]{0.49\textwidth}
        \includegraphics[width=\textwidth]{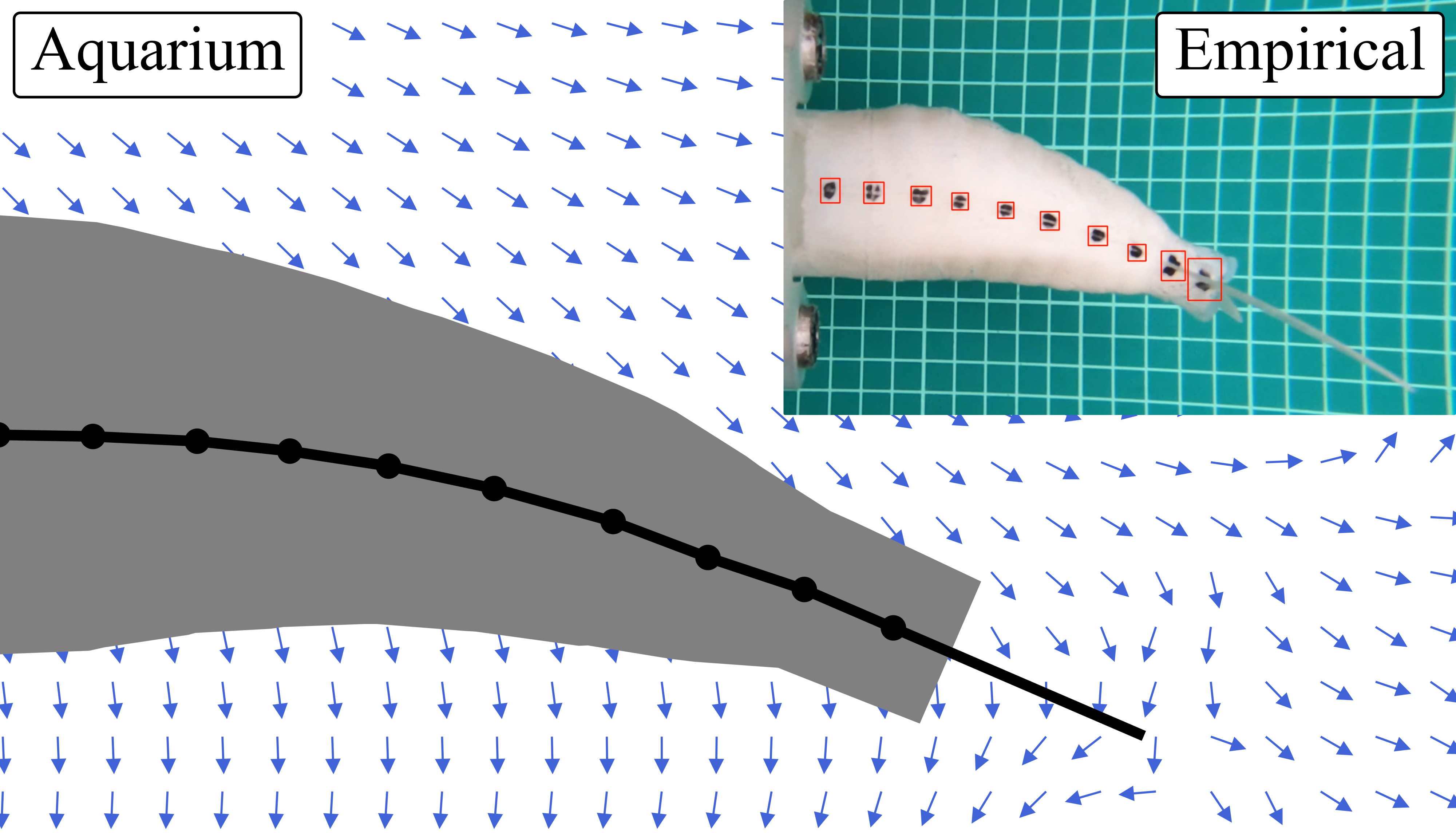}
        \caption{Aquarium simulation with corresponding hardware experiment. The tracked markers on hardware, modeled as joints of the multi-link, simulated representation, are boxed in \color{red}red \color{black} with corresponding center-line links and joints in simulation shown in black.}
        \label{fig:soft-tail-hardware-vs-sim}
    \end{subfigure}
    \begin{subfigure}[t]{0.49\textwidth}
        \includegraphics[width=\textwidth, height=0.57\textwidth]{hardware_validation.tikz}
        \caption{Time history of normalized thrust force generated by soft robotic tail from both simulation (black) and empirical (\color{red}red\color{black}) results.}
        \label{fig:soft-tail-force-histories}
    \end{subfigure}
    \caption{Unsteady fluid-structure interaction of a fixed-base soft robotic fish tail with matching hardware demonstration. The robotic fish tail starts at rest in initially still water before being actuated at 3Hz. Aquarium is able to properly simulate the transient flow and resulting forcing effects that are also observed empirically, especially the phase and frequency.}
    \vspace{-\baselineskip}
\end{figure*}

\section{EXPERIMENTAL RESULTS} \label{sec:RESULTS}

This section presents the results of several simulation experiments to evaluate the FSI physics of Aquarium with comparisons to both other numerical works and hardware experiments. This includes the classic cylinder-in-free-stream benchmark and a real-to-sim demonstration of a real-world soft robotic fish tail. We also demonstrate the full differentiability of Aquarium with a gradient-based optimization example that involves maximizing the thrust of an oscillating diamond foil. These examples, along with the open-source implementation of Aquarium are available at: \url{https://github.com/RoboticExplorationLab/Aquarium.jl}

\subsection{Cylinder in Free Stream}

We simulate the classic benchmark example of a cylinder in free-stream conditions as shown in Figure \ref{fig:cylinder_setup}. To do so, we define the simulation environment to have inflow and outflow boundary conditions on the left and right boundaries, respectively. The inflow boundary condition is defined to be the free-stream velocity, $u_{\infty}$, while the outflow boundary condition allows vortices to exit freely. We define the top and bottom fluid boundaries to have far-field conditions (i.e., the same velocity as inflow) to also simulate free-stream conditions when sufficiently distanced from the cylinder.

To study Aquarium's generalizability to varying Reynolds numbers, we evaluate the resulting steady-state behaviors under various free-stream velocities. As seen in Figure \ref{fig:freestream-cylinder-vorticity}, Aquarium is able to capture both the steady-state vortex pairs at lower Reynolds numbers ($Re=40$) and the Kármán vortex street at higher Reynolds numbers ($Re=100$).

To study the flow-induced forces on the cylinder, we also evaluate the resulting steady-state drag and lift coefficients. As seen in Tables \ref{tab:results-cylinder-Re40} and \ref{tab:results-cylinder-Re100}, there is good agreement between Aquarium and previous numerical and empirical studies. This is also true for the non-dimensional Strouhal number, which characterizes periodic vortex shedding in the wake and is critical for studying bio-inspired swimming~\cite{triantafyllou_hydrodynamics_2000}.

\subsection{Soft Robotic Fish Tail}

To demonstrate Aquarium's ability to generalize beyond simple geometry in steady-state flow, we also simulate the periodic flapping of a soft robotic fish tail in initially still water, and validate it against a hardware experiment as shown in Figure \ref{fig:soft-tail-hardware-vs-sim}. The soft robotic fish tail is fabricated as described in \cite{zhang_learning_2021, zhang_creation_2022}, and has a left and right chamber that are pneumatically actuated with \SI{500}{\milli\bar} at \SI{3}{\hertz}. The hardware experiment involves fixing the tail to a force sensor, which collects a time history of the net thrust force while the tail is actuated. We use previously collected video and force measurement data from \cite{zhang_learning_2021}.

\begin{table}[h]
    \vspace{0.5\baselineskip}
    \caption{Steady-state results at $Re=40$}
    \vspace{-0.1\baselineskip}
    \begin{center}
    \begin{tabular}{lc}\toprule
         & \textbf{Drag Coeff.}\\\midrule
        Tritton \cite{tritton_experiments_1959} (\emph{empirical}) & 1.65\\
        Taira et. al \cite{taira_immersed_2007} & 1.55\\
        Ren et. al \cite{ren_stream_2012} & 1.57\\
        \textbf{Aquarium} & 1.75\\\bottomrule
    \end{tabular}
    \end{center}
    \label{tab:results-cylinder-Re40}
    \vspace{-\baselineskip}
\end{table}
\begin{table}[h]
    \caption{Steady-state results at $Re=100$}
    \vspace{-0.1\baselineskip}
    \begin{center}
    \begin{tabular}{lccc}\toprule
         & \textbf{Drag Coeff.} & \textbf{Lift Coeff.} & \textbf{Strouhal} \# \\\midrule
        Braza et. al \cite{braza_numerical_1986} & $1.325 \pm 0.008$ & $\pm 0.280$ & $0.164$ \\
        Ren et. al \cite{ren_stream_2012} & $1.335 \pm 0.011$ & $\pm 0.356$ & $0.164$ \\
        \textbf{Aquarium} & $1.481 \pm 0.010$ & $\pm 0.362$ & $0.174$ \\\bottomrule
    \end{tabular}
    \end{center}
    \label{tab:results-cylinder-Re100}
    \vspace{-1.5\baselineskip}
\end{table}

In simulation, we approximate the soft robotic fish tail as a ten-link, serial-chain, rigid-body model with the joints located at corresponding marker positions along the center line of the robot, as shown in Figure \ref{fig:soft-tail-hardware-vs-sim}. The body profiles of each link are approximated by a linear interpolation between the respective joint widths, and the fin is represented as a 1D link. The simulated motion is prescribed using a forward-kinematics model, where the joint angles are determined using CSRT \cite{lukezic_discriminative_2017} marker tracking from a pre-recorded video of the hardware experiment. The fluid environment is also modeled to recreate the hardware experiment: a $\SI{0.6}{\meter} \times \SI{0.6}{\meter}$ cavity with wall-like boundary conditions ($u_{\infty}=0$) filled with initially still water. Simulated boundary pressure forces, $f^{b}$, are integrated over the boundary and compared to the empirical force-sensor measurements. To best match our 2D simulation to the 3D hardware experiment, we normalize both datasets' maximum thrust values to one.

As seen in Figure \ref{fig:soft-tail-force-histories}, there is good agreement between the phase and frequency of the normalized thrust forces with respect to time, demonstrating the simulation's ability to capture transient-flow effects on a moving boundary. Potential sources of error include Aquarium's inability to capture the full 3D effects (i.e., flow over the 3D contour of the fish tail) present in the experiment. This is a limitation of a 2D simulation, and future work is planned to extend Aquarium to 3D. Other potential sources of error include hardware fabrication error as well as the geometric and kinematic approximations of the multi-link representation.

\subsection{Optimization using Aquarium Gradients}

\begin{figure}[t]
    \vspace{0.5\baselineskip}
    \centering
    \includegraphics[height=4.5cm]{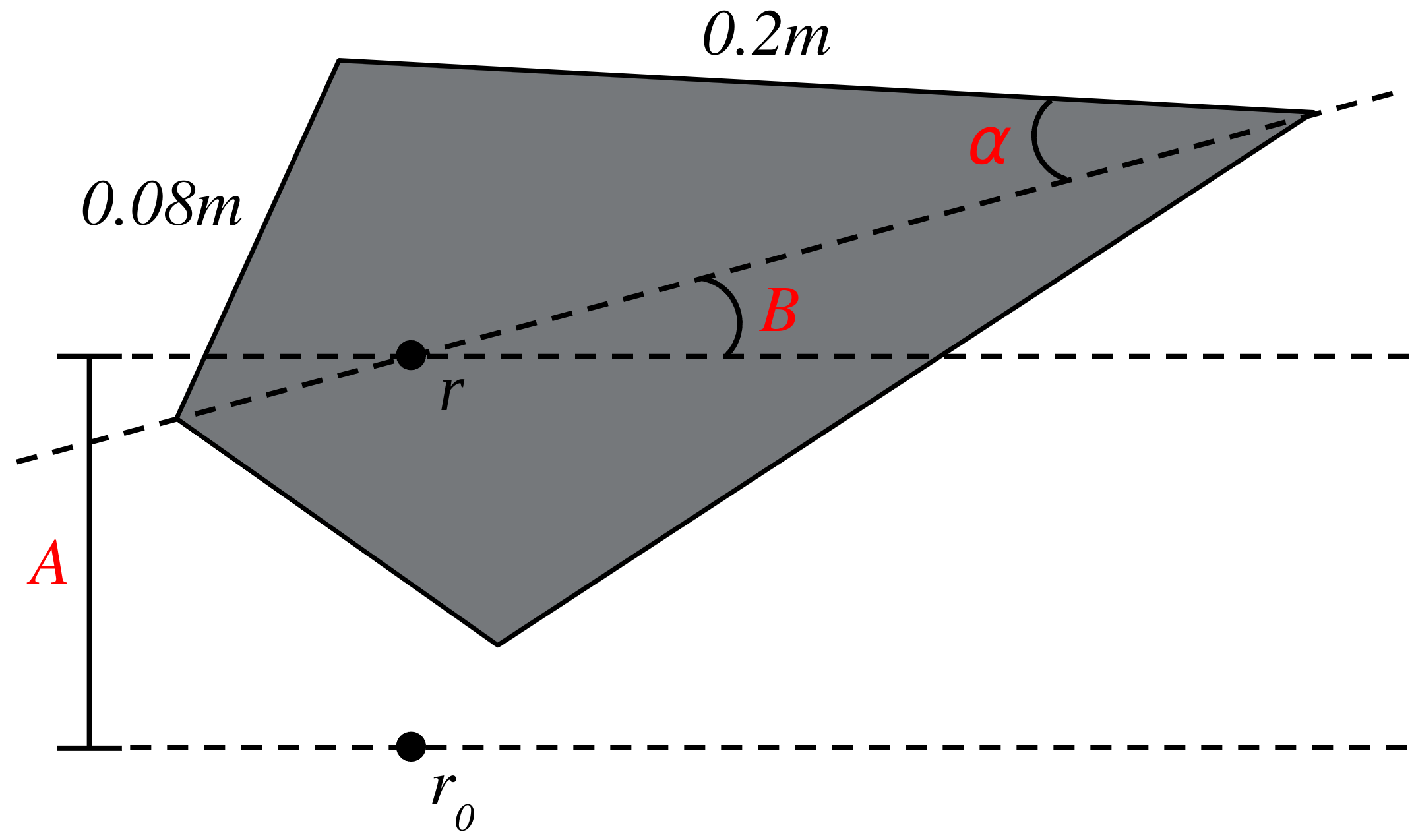}
    \caption{Oscillating diamond-foil optimization setup with decision variables (1 shape and 3 gait parameters) shown in \color{red}red\color{black}. \color{red}$\alpha$ \color{black} is the angle varying the foil thickness, \color{red}$A$ \color{black} is the heave amplitude, and \color{red}$B$ \color{black} is the pitch amplitude. The relative pitch-heave phase, $\phi$ (not shown) is also a decision variable.}
    \label{fig:diamond_foil_setup}
    \vspace{-\baselineskip}
\end{figure}

To showcase the full differentiability of Aquarium, we perform a shape-and-gait co-optimization of an oscillating diamond foil in free-stream conditions using the gradient-based, limited-memory BFGS (L-BFGS) algorithm~\cite{nocedal_numerical_2006} as shown in Figure \ref{fig:foil-optimization}. Specifically, we aim to maximize the thrust (i.e., minimize drag) generated by the foil and formulate the objective as the integral of the thrust force over an oscillation period. The optimization is performed over a range of chord-wise $Re \in [620, 828]$. This demonstrates Aquarium's ability to be used in optimization in unsteady flow environments. We represent the diamond foil's shape with constant edge lengths $(\SI{0.08}{\meter}, \SI{0.2}{\meter})$ and an angle parameter $\alpha$ that determines the foil thickness as seen in Figure \ref{fig:diamond_foil_setup}. Also seen in Figure \ref{fig:diamond_foil_setup}, the gait is determined by the heave amplitude $A$, pitch angle amplitude $B$, and relative pitch-heave phase $\phi$. To avoid degenerate cases, we impose box constraints on the decision variables with $6^{\circ} \leq \alpha \leq 23^{\circ}$, $\SI{0.05}{\meter} \leq A \leq \SI{0.15}{\meter}$, $0^{\circ} \leq B \leq 60^{\circ}$, and $0^{\circ} \leq \phi \leq 180^{\circ}$.

\begin{figure}[t]
    \vspace{0.5\baselineskip}
    \centering
    \includegraphics[width=0.475\textwidth, height=4.5cm]{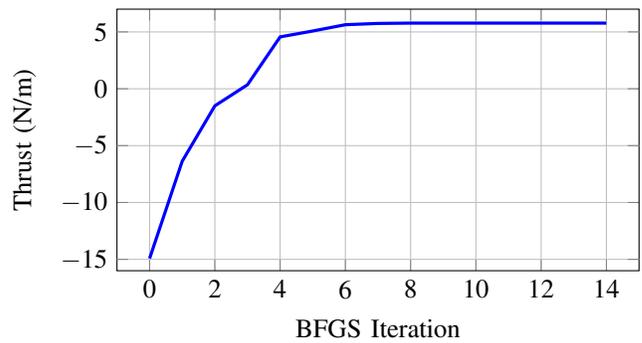}
    \caption{The reward (time-averaged thrust) throughout the co-shape-gait optimization of an oscillating diamond foil. Using gradients provided by Aquarium, the L-BFGS algorithm quickly converges to the near-optimized solution in 7 iterations with full convergence at 14 iterations.}
    \label{fig:BFGS-reward-iteration-history}
    \vspace{-\baselineskip}
\end{figure}

Using gradients provided by Aquarium, L-BFGS achieves a $139\%$ improvement in thrust after converging to a thrust-generating solution from a drag-inducing initial geometry and gait as seen in Figure \ref{fig:optimization-thrust-history}. Specifically, the foil thickness is minimized ($\alpha = 6^{\circ}$) and the heave amplitude is maximized ($A = \SI{0.15}{\meter}$) with a converged pitch and phase of $B \approx 27^{\circ}$ and $\phi \approx 74^{\circ}$ as seen in Figure \ref{fig:optimization-shape-progression}. Despite the differences in $Re$ regime and overall foil shape, there is still good agreement between this optimized gait profile to the empirical findings from Anderson et. al.~\cite{anderson_oscillating_1998}. In addition, L-BFGS was able to quickly converge to the optimized solution in only 7 iterations as seen in Figure \ref{fig:BFGS-reward-iteration-history}, showcasing the sample-efficiency achieved by gradient-based optimization with Aquarium.

The entire optimization process had a total runtime of $6$ hours on a $64$-core AMD Threadripper CPU. Each simulation rollout is performed over 100 time steps over a $300 \times 300$ fluid grid, with the KKT system in \eqref{eq:FSI-KKT} and \eqref{eq:gradient-KKT-system} solved using MKL Pardiso~\cite{schenk_pardiso_2001}. Aquarium is currently not optimized for computational performance, and we expect an order of magnitude improvement by solving the KKT system with a specialized sparse linear solver and efficiently implementing the adjoint method.

\section{CONCLUSIONS} \label{sec:CONCLUSION}

We have presented Aquarium, a fully differentiable fluid-structure interaction simulator that provides full, analytical gradients while accurately simulating coupled fluid and rigid-body dynamics in 2D. Aquarium improves on existing fluid simulators by offering three key features: 1) full differentiability with analytical gradients, which enables optimization in unsteady flow; 2) accurate and stable modeling of fluid dynamics by applying fully implicit integration to the full Navier-Stokes equations; and 3) explicitly formulated fluid-structure interaction that couples fluid physics with rigid-body dynamics. Aquarium enables a variety of optimization tasks --- including gait optimization, reinforcement learning, and hardware-controller co-design --- suited to robotics applications, where efficient locomotion may need to consider detailed flow physics both steady and unsteady.

In future work, we plan to address the current limitations of Aquarium, which include the lack of 3D simulation and soft bodies. Doing so will improve sim-to-real transfer over a wide-range of robotic systems while offering optimization over 3D geometries and gaits. We also plan to improve Aquarium's computational efficiency by implementing the adjoint method and a specialized sparse linear solver, making Aquarium more suitable for 3D simulation and high-dimensional optimization tasks such as the training of deep-learning models.

\newpage
\addtolength{\textheight}{-6cm}   
\bibliography{Aquarium.bib}
                                  
\end{document}